\documentclass{article}
\usepackage{arxiv}
\usepackage[utf8]{inputenc} 
\usepackage[T1]{fontenc}    
\usepackage{hyperref}       
\usepackage{url}            
\usepackage{booktabs}       
\usepackage{amsfonts}       
\usepackage{nicefrac}       
\usepackage{microtype}      
\usepackage{graphicx}
\usepackage{doi}
\usepackage{mathtools}
\usepackage{bm} 
\usepackage{siunitx} 
\usepackage{dsfont}

\usepackage{amssymb}
\usepackage{amsmath}
\usepackage{ulem}
\usepackage{xcolor}

\author{ 
{
        \hspace{1mm} Moritz Herrmann}\thanks{Corresponding author 
        } \\
    \footnotesize{\texttt{moritz.herrmann@stat.uni-muenchen.de}}\\
	Department of Statistics\\
	Ludwig-Maximilians-University\\
	Munich, Germany \\
	\And
	{
	\hspace{1mm}Fabian Scheipl} \\
	\footnotesize{\texttt{fabian.scheipl@stat.uni-muenchen.de}}\\
	Department of Statistics\\
	Ludwig-Maximilians-University\\
	Munich, Germany \\
    \\
}

\date{}

\renewcommand{\headeright}{PREPRINT}
\renewcommand{\undertitle}{PREPRINT}
\renewcommand{\shorttitle}{Unsupervised Functional Data Analysis}

\hypersetup{
pdftitle={Unsupervised Functional Data Analysis via Nonlinear Dimension Reduction},
pdfauthor={Moritz Herrmann, Fabian Scheipl},
pdfkeywords={Dimension reduction, Functional data analysis, Manifold methods, Unsupervised learning},
}

\newcommand{\hdspace}{\mathcal{H}}
\newcommand{\eucspace}{\mathbb{R}^D}
\newcommand{\funspace}{\mathcal{F}}
\newcommand{\embedspace}{\mathcal{Y}}
\newcommand{\pspace}{\Theta}
\newcommand{\mani}{\mathcal{M}}

\newcommand{\lcmcsize}{g}
\newcommand{\qlocal}{$\text{Q}^m_\text{local}$}

\newcommand{\imap}{ISOMAP}
\newcommand{\umap}{UMAP}
\newcommand{\dmap}{DIFFMAP}
\newcommand{\tsne}{t-SNE}
\newcommand{\mds}{MDS}
\newcommand{\AUC}{$AUC^m_{R_{NX}}$}
\newcommand{\AUCe}{$AUC^{dir}_{R_{NX}}$}
\newcommand{\AUCg}{$AUC^{geo}_{R_{NX}}$}
\newcommand{\ltwo}{$L_2$}
\newcommand{\metric}{$m$}
\newcommand{\lset}{a1-l, p1-l, c1-l, a2-l, p2-l, i2-l}
\newcommand{\nlset}{a2-sr, a3-hx, a3-sr, a3-sc, a3-tp}
\newcommand{\dir}{direct}
\newcommand{\dirs}{dir}

\begin{document}

\title{Unsupervised Functional Data Analysis via Nonlinear Dimension Reduction}
%




%
\maketitle              

\begin{abstract}
In recent years, manifold methods have moved into focus as tools for dimension reduction. Assuming that the high-dimensional data actually lie on or close to a low-dimensional nonlinear manifold, these methods have shown convincing results in several settings. This manifold assumption is often reasonable for functional data, i.e., data representing continuously observed functions, as well. However, the performance of manifold methods recently proposed for tabular or image data has not been systematically assessed in the case of functional data yet. 
Moreover, it is unclear how to evaluate the quality of learned embeddings that do not yield invertible mappings, since the reconstruction error cannot be used as a performance measure for such representations. In this work, we describe and investigate the specific challenges for nonlinear dimension reduction posed by the functional data setting.
The contributions of the paper are three-fold: First of all, we define a theoretical framework which allows to systematically assess specific challenges that arise in the functional data context, transfer several nonlinear dimension reduction methods for tabular and image data to functional data, and show that manifold methods can be used successfully in this setting. Secondly, we subject performance assessment and tuning strategies to a thorough and systematic evaluation based on several different functional data settings and point out some previously undescribed weaknesses and pitfalls which can jeopardize reliable judgment of embedding quality. Thirdly, we propose a nuanced approach to make trustworthy decisions for or against competing nonconforming embeddings more objectively.
\end{abstract}

\section{Introduction}

The ever growing amount of easily available high-dimensional data has led to an increasing interest in methods for dimension reduction in several contexts, for example image processing \cite{gong_intrinsic_2019} and single cell data \cite{becht_dimensionality_2019, kobak_art_2019}. Next to standard dimension reduction methods such as Principal Component Analysis (PCA) and Multidimensional Scaling (MDS), manifold methods have moved into focus in recent years. If the assumption that high-dimensional data actually lie on or close to a lower-dimensional Riemannian manifold holds, i.e., if the data have low \textit{intrinsic} dimension, nonlinear dimension reduction methods are often capable of detecting this intrinsic low-dimensional structure even if standard, in particular linear, methods fail to do so. In this paper, we describe and assess a general approach for extending established and state of the art manifold methods \imap~\cite{tenenbaum_global_2000}, \dmap~\cite{coifman_diffusion_2006}, \tsne~\cite{maaten_visualizing_2008}, and \umap~\cite{mcinnes_umap_2018} to functional data and use MDS as a default reference method for benchmarking.\\
\indent Functional data analysis (FDA) \cite[e.g.]{ramsay_functional_2005, wang_fdareview_2016}, which is an active field of research in statistics with many close connections to time series analysis, focuses on data in which the units of observation are realizations of stochastic processes over compact domains. 
This kind of data is another data type for which the manifold assumption is often reasonable: On the one hand, such data is infinite dimensional in theory and typically very high-dimensional in practice -- functional observations are usually recorded on fine and dense grids: For example, spectroscopic measurements are typically evaluated on thousands of electromagnetic wavelengths or electrocardiograms, measured at 100 Hz for 10 minutes, would yield 60,000 grid points each.
On the other hand, such signals typically contain a lot of structure, and it is often reasonable to assume that only few modes of variation suffice to describe most of the information contained in the data, i.e., such functional data often have low intrinsic dimension, at least approximately.\\
\indent An important complication is that FDA often faces the challenge of two kinds of variation, both of which can be of major interest: amplitude (i.e., ``vertical'') variation affecting the slope, level, and size of local extrema of a function and phase (i.e., ``horizontal'') variation affecting the location of extrema and inflection points. Phase variation, which can be conceptualized as elastic deformations of the domain of the functional observations, often results in complex nonlinear intrinsic structure \cite{chen_manifoldFDA_2012}. As our results show, this is true even for fairly simple phase variation structures. 
Despite some prior work \cite{chen_manifoldFDA_2012, dimeglio_template_2014} showing that manifold methods for functional data -- specifically, functional versions of \imap~\cite{tenenbaum_global_2000} -- can successfully deal with structured phase variation and yield efficient and compact low-dimensional representations and despite recent substantial progress in the development and application of manifold methods to tabular, image and video data \cite[e.g.]{kobak_art_2019, wang_image_2018}, manifold learning for functional data remains an underdeveloped topic. However, low-dimensional representations of functional data are highly relevant for real world problems. Finding reliable low-dimensional -- especially 2- or 3-dimensional -- representations of data is beneficial for visualization, description and exploration purposes in general. In FDA settings, this is especially crucial as the visualization of large data sets of functional observations is particularly challenging and quickly overwhelms analysts with ostensible complexity even if the underlying structures are actually fairly low-dimensional and simple, c.f Figure \ref{fig:ex-funs}. Moreover, finding informative low-dimensional representations of functional data is an essential preprocessing step for functional data, since these representations can be used as feature vectors in supervised machine learning algorithms which require tabular, not functional data inputs \cite{pfisterer_benchmarking_2019}.\\
\indent In this work, we thoroughly assess if manifold methods can be used to embed functional data, perform a careful evaluation of hyper-parameter tuning approaches for functional manifold methods and investigate the suitability of the derived embeddings in various settings. 
Specifically, we address the following questions:\newline (1) Are the manifold methods under investigation able to detect low-dimensional manifold structure of functional data? Special attention is given to assess the effects of phase variation.\newline 
(2) To what extent can automatic tuning strategies replace laborious and subjective visual inspection in order to obtain reliable embeddings in unsupervised FDA settings?

The remainder of the paper is structured as follows: In section \ref{sec:background} we specify notation and the theoretical framework, give a description of the embedding methods used and an overview over performance assessment and tuning approaches. Moreover, we motivate our study design. In section \ref{sec:simulation-study} we describe the design of the synthetic data simulations and assess the tuning and embedding approaches in these settings, in which the ``ground truth'' is available for verification. The concepts and insights developed on synthetic data are then brought to bear on three real data sets in section \ref{sec:real-data}. The findings of the study are finally discussed in section \ref{sec:discussion}.

\section{Background}\label{sec:background}
\subsection{Problem specification and framework}\label{sec:background:notation}

Nonlinear dimension reduction (NDR) \cite[e.g.]{bengio_representation_2014,lee_nonlinear_2007} is based on the assumption that high-dimensional data observed in a $D$-dimensional space $\hdspace$ actually lie on or close to a $d$-dimensional manifold $\mani \subset \hdspace$, with $d < D$ \cite{cayton_algorithms_2005}. One is then interested in finding an embedding $e:\hdspace \to \embedspace$ from the high-dimensional space to a low-dimensional embedding space $\embedspace$ such that $\embedspace$ is as similar to $\mani$ as possible. 

In most NDR applications, one simply considers $\hdspace = \eucspace$.
In a functional data setting, the situation is more involved: We define a d-dimensional \textit{parameter space} $\pspace \subset \mathbb{R}^d$ while $\funspace = \mathcal{L}^2(\mathcal{T})$, the space of square integrable functions over the domain $\mathcal{T}$, takes on the role of $\hdspace$, and $\phi: \pspace \to \funspace$ is a mapping from the parameter space to the \textit{function space}. We then observe functions $x_i(t) \in \funspace$ with $x_i(t) = \phi(\theta_i)$, which can have a complex but intrinsic low dimensional structure, depending on both the structure and dimensionality of $\pspace$ and the complexity of $\phi$.

\begin{figure}
    \includegraphics[width=\textwidth]{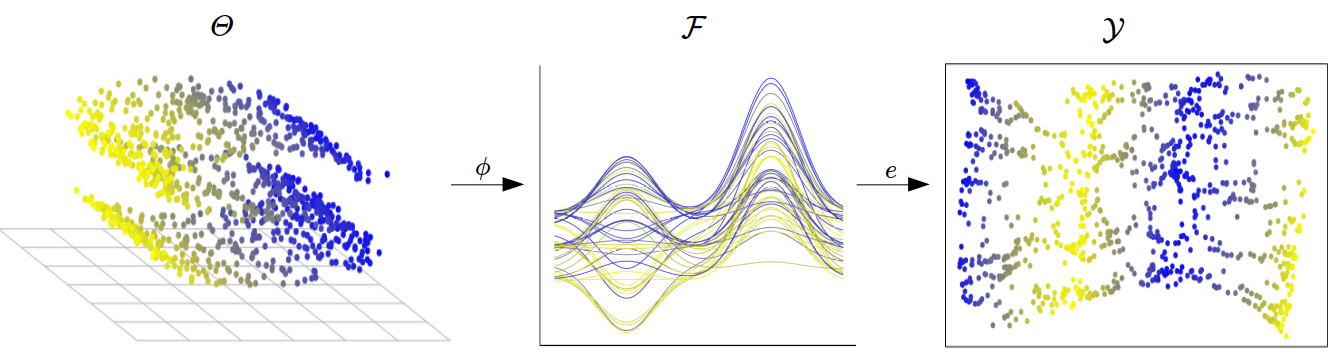}
    \caption{Framework for nonlinear dimension reduction of functional data.} \label{fig:concept}
\end{figure}

Transferring this to the NDR terminology, $\mani = \pspace$, i.e. the low-dimensional manifold is the parameter space. However, the observed data are functions in the subspace $\mani_{\funspace} \subset \funspace$, i.e., using the terms of \cite{chen_manifoldFDA_2012}, a functional manifold. Thus, using manifold methods, an embedding $e:\mani_{\funspace} \to \embedspace$ can be constructed. Specifically, that means we have the mappings \mbox{$\Theta \stackrel{\phi}{\to} \mathcal{\mani_{\funspace}} \stackrel{e}{\to} \mathcal{Y}$}, but only $e:\mathcal{F} \to \mathcal{Y}$ can be learned from the data (s. Figure \ref{fig:concept}). The question we try to answer is this: how well can the underlying global structure of $\pspace$ be recovered in an embedding learned from data on $\mani_{\funspace}$?

In general, however, it is not straightforward to define what ``recovering well" is supposed to mean in a specific NDR setting \cite{goldberg2008manifold}, a rarely discussed but crucial aspect.
In particular, the fact that manifolds which are locally homeomorphic to $\mathbb{R}^d$ by definition need not be homeomorphic to $\mathbb{R}^d$ globally needs to be taken into account.
E.g., a 2-sphere, although locally homeomorphic to the plane $\mathbb{R}^2$, can not be embedded into $\mathbb{R}^2$ globally by a single embedding $e$ without distortions. In differential geometry terms, only specific manifolds $\mathcal{M}$ like the famous ``Swiss roll'' can be represented by a single chart. However, since learning an embedding function is roughly equivalent to estimating a chart of the data manifold, manifold learning faces an ill-posed problem if the atlas of a data manifold requires more than one chart. 

Since it is not known in practice whether a single chart is sufficient or not, assessment of the embeddings achieved by manifold methods must 
always be considered under both local and global perspectives: (1) successful on a local level if the embedding $e:\hdspace \to \mathbb{R}^d$ yields an embedding space $\mathcal{Y}$ in which local structures are similar to local structures in $\mathcal{M} \subset \hdspace$, i.e., if $e$ preserves neighborhoods of (small) membership size $g$, and (2) globally successful if $\embedspace$ is as close to isomorphic to $\mathcal{M}$ as possible, e.g., if an underlying ``Swiss roll'' manifold is ``unrolled'' into a plane. Note that the latter is not possible for every manifold. 
Thus, we consider a learned embedding to be successful if the resulting configurations of data units in $\embedspace$ are as similar to the corresponding configurations in $\pspace$ as possible in the following sense:
In the conducted simulation study, we use simple parameter manifolds $\Theta$ which are mostly homeomorphic to $\mathbb{R}^d$ with $d \in \{1,2\}$, and -- in one case -- homeomorphic to the circle. This allows us to evaluate embedding methods and tuning approaches for the functional data generated from  $\Theta$ from a \textit{global} perspective, both by visual inspection and quantitatively. \textit{Local} characteristics are additionally investigated for the real data sets. 

As phase variation typically transforms the domain non-linearly, phase-varying functional data is very likely to live on a non-simple functional manifold $\mani_{\funspace}$ that is no longer globally isomorphic to $\mathbb{R}^d$ even if the generating parameter manifold $\Theta$ is a simple linear subspace, an issue leading to additional complexity in the FDA setting. 
By restricting our simulation study in part to fairly simple, linear $\Theta$, we are able to assess these non-obvious and previously undescribed effects of domain warping on the embeddings.

\subsection{Assessing performance of manifold methods}\label{sec:background:performace}
Since the methods we employ do not yield invertible embeddings $e$, we cannot evaluate them based on their reconstruction error $E[L(x, e^{-1}(y))]$, where $L$  is some loss function measuring the divergence between $x$ and its reconstruction $e^{-1}(y)$, which would objectively quantify the quality of an embedding in terms of the fidelity of its low-dimensional compression to the original data. Assessing embeddings qualitatively by visual inspection, as is widely done \cite[e.g.]{cayton_algorithms_2005,alaiz_diusion_2015,mcinnes_umap_2018},
cannot be automated, and so it does not scale to large-scale comparison and benchmark studies nor to the important task of tuning a method's hyperparameters.

Instead, several surrogate measures have been developed, often based on 
comparing the ranks of pairwise distances
between the high-dimensional data space and the learned embedding space \cite[e.g.]{venna_neighborhood_2001, chen_local_2009, lee_quality_2009, liang_new_2020}. We employ measures based on the normalized local continuity meta-criterion (LCMC) \cite{lee_type_2013}, since they are parameter free, yield a single scalar value -- a property particularly desirable if the measure is supposed to be used for tuning -- and allow to assess local and global performance.
The LCMC is based on the measure $$Q_{RX}(\lcmcsize) =\frac{1}{\lcmcsize} \underbrace{\frac{1}{n} \sum^n_{i=1}  \vert \mathcal{N}^{\hdspace}_\lcmcsize(i) \cap \mathcal{N}^{\embedspace}_\lcmcsize(i) \vert}_{N_\lcmcsize},$$ which quantifies the amount of overlap between the memberships of neighborhoods of a certain size in the two spaces \cite{lee_quality_2009, kraemer_dimred_2018}. The $\lcmcsize$-neighborhood $\mathcal{N}_\lcmcsize(i)$ is defined as the set of those $\lcmcsize$ objects which are closest to $i$ in the respective space according to a suitable distance measure, and $N_\lcmcsize$ measures the mean overlap obtained by averaging the cardinalities of the intersections between all such neighborhoods in the high dimensional space $\hdspace$ and the low dimensional embedding space $\embedspace$. The factor $\frac{1}{\lcmcsize}$ is a normalization factor. 
The normalized LCMC is then defined as $$R_{RX}(\lcmcsize) = \frac{(n-1)Q_{RX}(\lcmcsize)-\lcmcsize}{n-1-\lcmcsize},$$
which also accounts for random overlap \cite{chen_local_2009, lee_type_2013}. A value of 0 is expected for a random embedding, i.e., the agreement between $\lcmcsize$-neighborhoods in $\embedspace$ and in $\hdspace$ is the same as that of a random configuration of objects in $\embedspace$. A value of 1 indicates a perfect embedding with complete identity of all $\lcmcsize$-neighborhoods in the two spaces \cite{lee_type_2013, lee_quality_2009}. 

The choice of neighborhood size $\lcmcsize$, however, is crucial and has a strong influence on whether an embedding is judged to be successful or not. For large $\lcmcsize$, these metrics quantify the preservation of global structure in the embedding, for small $\lcmcsize$, the preservation of local structure. 

To circumvent the problems that come with the choice of $g$, one can compute parameter free measures based on $R_{NX}(\lcmcsize)$ \cite{kraemer_dimred_2018}. Regarding $R_{NX}(\lcmcsize)$ as a function of $\lcmcsize$ and averaging the function values on either side of its maximum at $g_{max}$, leads to both a local and a global performance measure: 
$$\text{Q}_\text{local} = \frac{1}{g_{max}}\sum_{g=1}^{g_{max}}Q_{RX}(\lcmcsize) \qquad \text{and} \qquad \text{Q}_\text{global} = \frac{1}{n - g_{max}}\sum_{g=g_{max}}^{n-1}Q_{RX}(\lcmcsize).$$
To quantify the overall performance of an embedding, the \textit{area under the $R_{NX}(\lcmcsize)$-curve} $$AUC_{R_{NX}} = \frac{\sum_{g=1}^{n-2}R_{NX}(g)}{\sum_{g=1}^{n-2}\frac{1}{g}}$$ can be computed \cite{kraemer_dimred_2018}. Given such a scalar measure of performance, embedding methods can then be tuned by maximizing this measure over the different hyperparameter settings.

However, while it is known that the choice of the distance metric has a strong influence on embedding methods \cite{Venna2010}, the influence of the distance metric used to compute the $\lcmcsize$-neighborhoods in the surrogate performance measures remains unclear. Since these measures are based on $\lcmcsize$-neighborhoods, the proximity metric used to define the neighborhoods in the respective spaces is crucial, especially if the goal of the analysis is to ``unfold'' the global structure of the manifold. 
In order to recover the global manifold structure, neighborhoods in the high-dimensional space should be defined using \textit{geodesic} distances rather than Euclidean distances, since only the former represent long-range distances on the manifold correctly, while the latter are merely distances in the ambient space. This distinction is likely to be highly relevant especially if the observed high-dimensional data manifold has a complex structure, such as $\mani_{\funspace}$ in our situation, and when the measures are supposed to be used for automatic parameter tuning for optimal recovery of global structure. 
In the following we use \AUC and \qlocal, where \metric~$\in \{\dirs, geo\}$~indicates the distance metric used to calculate the neighborhoods for performance assessment, i.e. in this study $\lcmcsize$-neighborhoods in  $R_{NX}(\lcmcsize)$ are computed either using \ltwo~distances (i.e. Euclidean distances in $\mathbb{R}^D$) or geodesic distances. In the following we use the term \textit{\dir} distance instead of \ltwo~distance to emphasis the conceptual difference of distance measures which merely quantify proximity in the ambient space (hence \textit{\dir} distance) and \textit{geodesic} distance measures quantifying proximity on a (nonlinear) manifold. Note, this is a general conceptual difference. Most standard distance metrics can be regarded as \dir~distance measures in the particular space and geodesic distances as computed here can be obtained based on several of these \dir~distance metrics.
We will show that direct distances such as the \ltwo~distance can yield very misleading results when used in the surrogate performance measures and that tuning approaches can thus lead to far form optimal configurations.

\subsection{Embedding methods and tuning approach}
In this study, we compare nonlinear dimension reduction methods \textit{isometric feature mapping} (\imap) \cite{tenenbaum_global_2000}, \textit{diffusion map} (\dmap) \cite{coifman_diffusion_2006}, \textit{t-distributed stochastic neighborhood embedding} (\tsne) \cite{maaten_visualizing_2008}, and \textit{uniform manifold approximation and unfolding} (\umap) \cite{mcinnes_umap_2018} for functional data. All these methods have \textit{locality parameters} that control whether (rather) local structures or (rather) global structures are considered, that is how much ``context" of the respective data points is taken into account while constructing the embedding. These parameters influence the result strongly and need to be tuned.  We apply MDS as a simple tuning-free benchmark reference method.

\imap~is based on classical MDS. In contrast to MDS it is capable of unfolding the intrinsic structure of a data set. The algorithm consist of three steps. First, a nearest-neighbor-graph is constructed based on a suitable direct distance metric, usually the \ltwo~metric. This requires defining a neighborhood size either by a distance threshold $\epsilon$ or by the number of neighbors $k$ to be included. This parameter is the main tuning parameter of the algorithm. In the next step, shortest-path or geodesic distances among all points are computed based on the nearest-neighbor graph. These distances are then supplied to classical MDS, which embeds the observations accordingly. \imap~is supposed to be particularly suited to detect global structures \cite{tenenbaum_global_2000}. \\
\dmap~is another spectral embedding method projecting on the eigenvectors of a diffusion operator on the data manifold. Proximity of data points is defined by a kernel function whose width acts as a tunable locality parameter \cite{coifman_diffusion_2006, ma2011manifold}.\\
\tsne, which has been state of the art for several years \cite{mcinnes_umap_2018}, builds upon stochastic neighborhood embedding (SNE). In contrast to the aforementioned methods, (t-)SNE transforms proximities between data points into conditional probabilities of them being neighbors in the respective space and then minimizes the KL divergence of the implied distribution in the original space from that in the embedding space. 
The perplexity of the implied distribution in the original space acts as a tunable locality parameter.
\\
\umap~\cite{mcinnes_umap_2018} is a state-of-the-art manifold learning method based on three assumptions -- uniformly distributed data on a locally connected manifold equipped with a locally constant metric. It computes a fuzzy topological representation of the manifold based on a nearest-neighbor-graph. The number of nearest neighbors serves as a tunable locality parameter. \\
\\
\indent In addition to the investigation of how successfully the intrinsic manifold structure of a functional data set can be detected and unfolded in general, we also want to investigate how reliably automatic tuning approaches identify suitable hyper-parameter settings. We consider the parameters steering the degree of locality as the main tuning parameter of these embedding methods. Using the terminology of the respective R packages, these are the neighborhood sizes \verb|k| for \imap, \verb|n_neighbors| for \umap, the \verb|perplexity| for \tsne~and \verb|eps.val| for \dmap. Hereinafter we refer to these parameters as \textit{locality parameters}. Moreover, the methods are not supplied with the raw data matrices, but with distance matrices instead. Note, this means that an initialization via PCA is not performed for \tsne~using \verb|Rtsne|.
To investigate these aspects, we compute both the \dir~and geodesic distance matrix for each simulated data set in the function space as well as in the parameter space. For a given parameter configuration, the \dir~distance matrix obtained from the function space is then input to the respective embedding method. Finally, \dir~distances of the learned embeddings are computed. 
As the performance measures are based on the comparison of $\lcmcsize$-neighborhoods in the high dimensional and the embedding spaces, we compute the performance measures with respect to both the parameter space as well as the function space based on \dir~and the geodesic distance neighborhoods in the simulation settings. Recall, \dir~distances represent distances in the ambient space rather than distances on the manifold and the resulting neighborhoods are thus unlikely to be well suited for performance assessment and tuning if the intrinsic structure is nonlinear, especially for larger neighborhood sizes.\\
\indent Parameters are tuned for optimal performance via an extensive grid search, c.f. Table \ref{tab:grids}. For \dmap~we compute a data set-specific starting value $\epsilon_s$ via \verb|epsilonCompute|, which is the default value of the method, and use this to define the search grid of the locality parameter. In the synthetic data settings of Section \ref{sec:simulation-study}, we can perform a ``ground truth''-based meta assessment, in which we evaluate the effect of \dir~and geodesic distances. Moreover, we can compare the results achieved by tuning based on performance measures computed in the function space with those achieved by a practically infeasible ``oracle'' tuning method that uses corresponding performance measures computed in the unobservable true parameter space instead.  

\begin{table}
    \caption{Parameters of the manifold methods which are subjected to tuning. The second column shows the total amount of different parameter configurations in the tuning parameter grid, the third column the locality parameter, the fifth column the embedding dimension parameter, and the last column further parameters tuned. The ``grid''-columns display the search grids of the parameters in the preceding column. The embedding dimension grid for \tsne~differs from the other grids because the implementation does not allow the embedding dimension to be greater than three.}
    \centering
    \resizebox{\textwidth}{!}{
    \begin{tabular}{|l|l|l|l|l|l|l|}
      \hline
      Method & \# & locality param. & grid & embedding dim. & grid & further param. \\
      \hline
      \mds  & 4    & - & - & \texttt{k} & [2, 5] &  - \\      
      \imap & 1300 & \texttt{k} & [3, 975] & \texttt{ndim} & [2, 5] & - \\
            & &         & step size: 3   &  &  &  \\
      \dmap & 6000 & \texttt{eps.val} & [$0.15 \epsilon_s$, $1.85 \epsilon_s$] & \texttt{neigen} & [2, 5] & \texttt{t} \\
            & & & length: 250  &  &  & \\
      \umap & 18720 & \texttt{n\_neighbors} & [5, 975] & \texttt{n\_components} & [2, 5] & \texttt{min\_dist} \\
            &  &  & step size: 5 & &  & \texttt{n\_epochs} \\
            &  &   &   &   &   & \texttt{init} \\
      \tsne & 21184 & \texttt{perplexity} & [3, 333] & \texttt{dims} & [2, 3] & \texttt{theta} \\
            & &  & step size: 1  &  &  & \texttt{max\_iter} \\
            & &  &   &  &  & \texttt{eta} \\
            & &  &   &  &  & \texttt{exaggeration} \\
      \hline
    \end{tabular}
    }
    \label{tab:grids}
\end{table}

\subsection{Study design}\label{sec:background:study-design}
Since we are interested in whether manifold methods and tuning approaches can be used for functional data sets in general, a thorough evaluation design is inevitable. For supervised learning algorithms a wide and comprehensive body of literature exists on the conduction of neutral and objective comparison and benchmark studies \cite[e.g.]{van2018benchmarking, boulesteix2013plea, eugster2011benchmark}.

How to reliably evaluate algorithms and meta learning approaches in unsupervised settings, however, is not as clear. Due to the lack of an outcome variable, clearly defined objectives to optimize against are usually not available. This makes the comparison of unsupervised learning algorithms in general, and the assessment of meta learning approaches such as tuning in particular, prone to overoptimistic findings. General frameworks for systematic benchmarks in this context are still in their infancy \cite{van2018benchmarking}. 

In particular, nonlinear dimension reduction and manifold learning are often confronted with the lack of a clearly defined objective in terms of the achieved reconstruction error if the methods do not yield an invertible mapping.
Since there is no standard benchmark procedure for unsupervised learning generally agreed upon, we devised the following procedure in order to avoid overoptimistic conclusions in our study as much as possible:

Based on the problem specification and the theoretical framework defined in Section \ref{sec:background:notation}, we first conduct a simulation study to assess embedding methods and the considered tuning approaches in settings where the ground truth is known. This allows us to investigate possible strengths, weaknesses and pitfalls in settings that allow for objective evaluations based on a known ``ground truth''.
We then apply the approach to real data sets where qualified assumptions about the intrinsic structure of the data can be made due to substantial considerations and analysis of previous studies. Although knowledge of the intrinsic structure is less certain than in the ground truth simulations, it is still possible to ``objectively" evaluate the embeddings, at least conditional on certain substantially justified assumptions, in these settings. To some extent, this lets us examine whether the insights obtained in simulated data settings also hold for real data. Finally, we compute embeddings for a real data set for which information about its intrinsic structure can not be justifiably inferred from prior substantial considerations -- i.e., a fully unsupervised problem. We will show that such a setting can pose severe problems due to nonconforming embeddings that are hard to tackle, but that ``principled'' choices between nonconforming embeddings may nevertheless be possible based on the insights gained from the simulation study and from data applications in which some prior knowledge is available.

\section{Simulation study}\label{sec:simulation-study}
Based on the framework described in Section \ref{sec:background:notation} we can systematically assess the utility of the described manifold methods for functional data settings by means of a simulation study. This allows to compare the embeddings to a ground truth which is essential in a unsupervised learning problem to come to reliable conclusions. In particular, we can assess the influence of some factors likely to lead to strongly nonlinear intrinsic structure of functional data manifolds, i.e., nonlinear domain warping (phase variation) and an underlying nonlinear parameter space $\pspace$, by systematically controlling these sources of variation.

\subsection{Experiment design}
Loosely speaking, the simulation design is based on two peaked functions derived from Gaussian pdfs over domain $[0, 1]$.  Variation is achieved by randomly changing the locations, widths and heights of the peaks, in total leading to eleven considered settings, six based on a linear parameter space, including three with nonlinear phase variation, and five based on a nonlinear parameter space and amplitude variation only. We can thus assess the effect of nonlinear phase variation and a nonlinear parameter space separately using the first six settings and the last five settings, respectively. 

Specifically, we consider the following functional manifold\newline $$\mani_{\funspace} = \left\{x \in \mathcal{L}^2([0, 1]): x(t) = \phi(\mathbf{\theta})\right\},$$
with $\theta = (\mathbf{a}, \mathbf{p})$ and $\phi(\mathbf{\theta}) = b(w(t; \mathbf{p}); \mathbf{a})$. Amplitude variation in $\mani_{\funspace}$ is parameterized as\newline
$$b(t; \mathbf{a}) = \frac{ a_1}{\sqrt{0.1\pi}} \{a_2 \cdot n(t, 0.25) + a_3 \cdot n(t, 0.75)\} + a_4,$$
with $n(t, \mu) = \exp\left(-\frac{(t - \mu)^2}{0.1}\right)$. Depending on the setting, phase variation is parameterized as the identity warping  $w(t; \mathbf{p}) = t$, a linear warping $$w(t; \mathbf{p}) = 
\begin{cases} 
    p_1t & \text{for } t \in [0, 0.5] \\ 
    (2-p_1)(t - 1) + 1 & \text{for } t \in (0.5, 1]\\
\end{cases},$$ a power warping $w(t; \mathbf{p}) = t^{p_2}$ or as $ w(t, \mathbf{p}) = \text{B}(t; p_3, p_4)$, where $B(\cdot; a, b)$ is the cdf of a Beta$(a,b)$ distribution.

The considered settings are obtained by selecting up to three of the parameters $a_1, a_2, a_3, a_4, p_1, p_2, p_3, p_4$, i.e., the considered settings have at most 3 degrees of freedom (df). Inactive parameters are set to constant values, e.g., $a_1 = 1$ and $a_4 = 0$. If no warping parameter is selected, the identity warping is applied. Varying both amplitude \textit{and} warping parameters in a setting induces joint amplitude and phase variation. Dependencies between amplitude and phase parameters induce dependencies between amplitude and phase variation.

\begin{figure}
    \includegraphics[width=\textwidth]{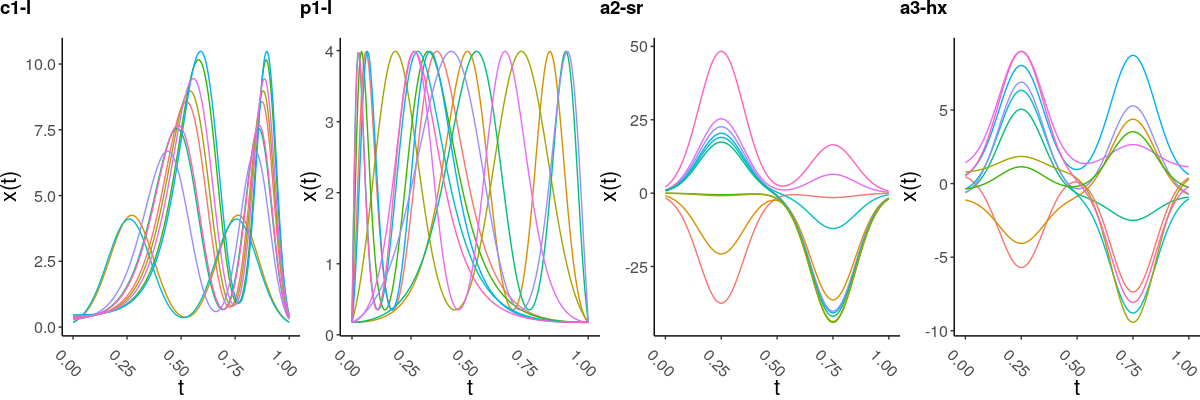}
    \caption{Example functions with 1 df coupled joint amplitude and phase variation (c1-l), 1 df phase variation (p1-l), and 2 and 3 df amplitude variation (a2-sr and a3-hx). c1-l and p1-l are based on linear, a2-sr and a3-hx are based on nonlinear parameter manifolds.} \label{fig:ex-funs}
\end{figure}

The active parameters are either drawn uniformly from a linear parameter space, i.e., the manifold $\Theta$ is a linear subspace, or from a nonlinear parameter space, i.e, a (nonlinear) manifold. Note that the power function and the beta cdf warping are nonlinear transformations of the domain, thus we obtain a non-linear functional data manifold even if the parameter space $\Theta$ is linear. In the linear case we let $a_i, p_j \sim U[0.5, 3]$, $i = \{1,...,4\}$ and $j = \{2, 3 ,4\}$, and $p_1 \sim U[0.01, 0.99]$. In the nonlinear case, parameter values are drawn uniformly from one of five different manifolds: the Swiss roll (1D and 2D), the 1D helix, the 2D S-curve, and the 2D two-peaked (tp) surface as provided by the R package \verb|dimRed| \cite{kraemer_dimred_2018}. For each setting, 1000 functional observations are generated based on 200 grid points. 
A summary of the simulation settings is provided in Table \ref{tab:sim_sets} and Figure \ref{fig:ex-funs} displays samples of functions with phase variation, amplitude variation, and joint, coupled amplitude and phase variation. 
\begin{table}
    \caption{Overview of simulation settings.}
    \centering
    \begin{tabular}{|l|l|l|l|l|l|l}
      \hline 
      setting & df & parameter space & variation & parameter & warping \\
      \hline
      a1-l & 1 & linear  & amplitude & $a_1$       & identity \\
      p1-l & 1 & linear  & phase     & $p_1$       & linear   \\
      c1-l & 1 & linear  & coupled & $a_1 = p_2$ & power      \\
      a2-l & 2 & linear  & amplitude & $a_2, a_3$  & identity \\
      p2-l & 2 & linear  & phase     & $p_1, p_2$  & beta cdf \\
      i2-l & 2 & linear  & independent & $a_1, p_1$  & power \\
      a2-sr & 2 & 1D Swiss roll & amplitude & $a_1, a_2$ & identity \\
      a3-hx & 3 & 1D helix & amplitude  & $a_1, a_2, a_3$ & identity \\
      a3-sr & 3 & 2D Swiss roll & amplitude & $a_2, a_3, a_4$ & identity \\
      a3-sc & 3 & 2D S-curve & amplitude & $a_2, a_3, a_4$ & identity \\
      a3-tp & 3 & 2D tp surface & amplitude & $a_2, a_3, a_4$ & identity \\
      \hline   
    \end{tabular}
    \label{tab:sim_sets}
\end{table}

\subsection{Results}
The results show that functional data can -- in principle, i.e. given the ground truth to compare against -- be embedded with methods developed primarily for images or tabular data. 

To start with we also sketch the effect of nonlinear warping in the next subsection. For that we consider settings \lset~with simple parameter spaces $\mathbb{R}^d$ first. Based on embeddings of the reference method MDS we analyze some specific pitfalls which result from the drawbacks of the performance measures described in Section \ref{sec:background:performace}. We finally turn to the settings with more complex parameter spaces, \nlset, additionally evaluating tuning approaches based on the surrogate performance measures. We show that based on the correct tuning approach, it is possible to (automatically) obtain high quality embeddings for these settings as well. 

\subsubsection{Embedding functional data with phase and amplitude variation}
\begin{figure}
    \includegraphics[width=\textwidth]{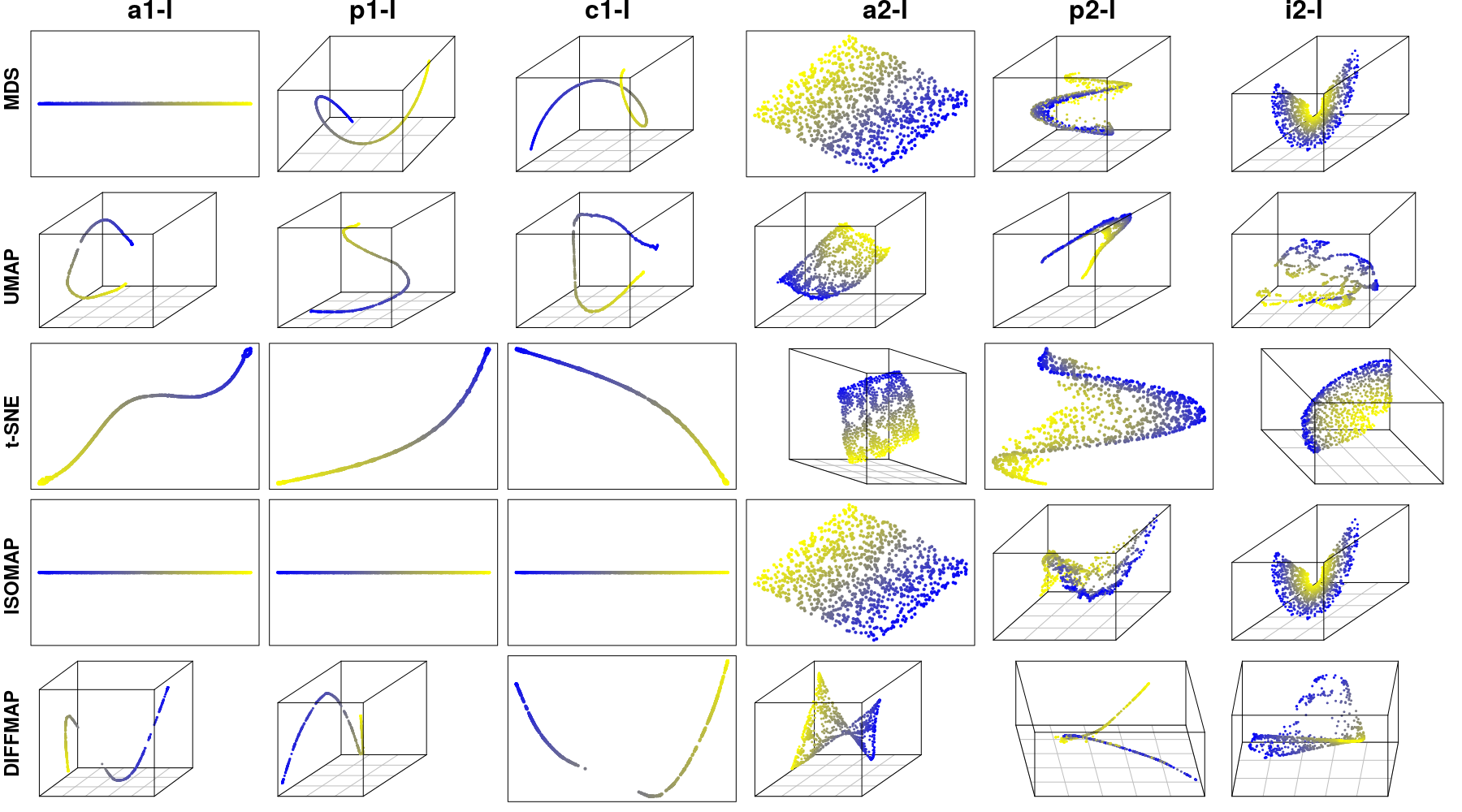}
    \caption{Embeddings for settings \lset. Color scale encodes value of the first parameter in $\pspace$. Since units for the embeddings are arbitrary, we omit axis labels here and in the following figures to save space.} \label{fig:pha-amp}
\end{figure}

In this section we highlight two essential aspects. First of all, the findings indicate that it is possible to successfully embed functional data using manifold methods, at least in these simple settings. In addition, we provide evidence that things can get rather complicated quickly if warping comes into play even if the underlying parameter space is a simple linear one. That said, the settings we consider here, \lset, include amplitude as well as phase variation and also both coupled and independent joint phase and amplitude variation. \\
\indent As can be seen in Figure \ref{fig:pha-amp}, \imap~is particularly successful. Clearly, perfect linear embeddings are achieved in settings with one degree of freedom and two degrees of freedom alike (a1-l, p1-l, c1-l, a2-l). Note that phase variation is induced by a nonlinear polynomial warping function in the setting with coupled phase and amplitude variation c1-l. Nevertheless, the functional manifold can be perfectly unfolded into its underlying linear structure. This is not the case for setting p2-l, where phase variation is induced by the Beta cdf. Here, the resulting structure of the functional manifold becomes more challenging to unfold into $\mathbb{R}^2$. 

This shows why embedding functional data can be especially complex and why it is important to use simple, low-dimensional parameter spaces for this study: the functional manifolds we are dealing with become nonlinear even though they are based on a deceptively simple parameter space. Moreover, consider setting i2-l, which has independent amplitude and phase variation (based on power warping). Even though the data are embedded nearly linearly, the distribution of the parameter values of the amplitude variation, indicated by colour code, no longer follows a simple linear direction in the embedding space. This also indicates the more complex structure of the functional manifold induced by the power warping. 

For the embeddings of the other methods similar findings can be reported, however overall they are not as successful as the \imap~embeddings. In the 1-dimensional settings most embeddings are not perfectly linear. Moreover, unlike \tsne, \umap~and \dmap~fail to recover setting i2-l respectively i2-l, p2-l, and a2-l. 
To sum up, manifold methods apparently seem to be able to recover the manifold structure of functional data, but how successfully they do so strongly depends of the complexity induced by the transformation of the parameter space. Structured phase variation can quickly lead to very complex functional manifolds which may not be embeddable faithfully in the sense that the functional data manifold is no longer isomorphic to a low-dimensional linear subspace even if it is based on a very simple linear parameter space.

\subsubsection{Pitfalls of surrogate performance measures}\label{sec:measures:pitfalls}
\begin{figure}
    \includegraphics[width=\textwidth]{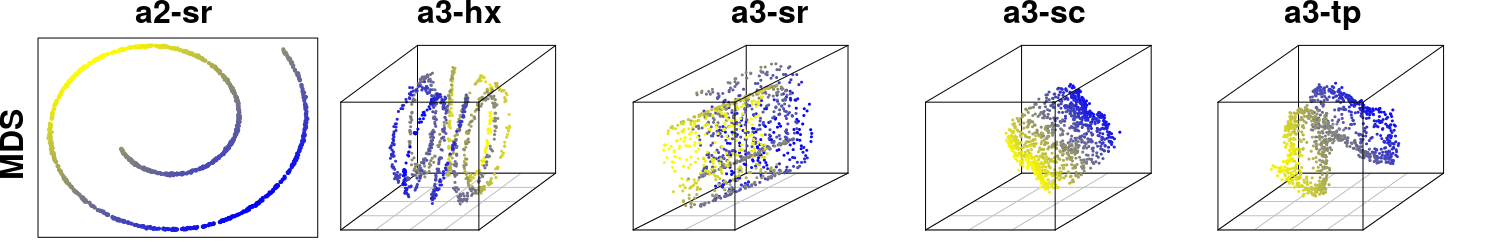}
    \caption{MDS embeddings for settings with nonlinear parameter space. Color scale encodes value of the first parameter in $\pspace$} \label{fig:embs-mds-nl}
\end{figure}

As outlined in section \ref{sec:background:performace}, the proposed surrogate performance measures suffer from some drawbacks. Here we show that one of the most worrisome resulting pitfalls is that these measures can frequently indicate high performance values even if the embedding is not of high quality in terms of ground-truth performance in the underlying parameter space. \\
\indent To demonstrate the issue we concentrate on \mds~embeddings of the nonlinear settings \nlset. As can be seen in Figure \ref{fig:embs-mds-nl}, simple \mds~is not able to \textit{unfold} the functional manifolds into embeddings on linear subspaces or the circle. However, assessing the embedding, e.g. of setting a3-hx, using \AUCe based on direct $L_2$-distance neighborhoods in the function space $\funspace$, i.e. the standard way of calculating distances, would indicate a perfect embedding quality of 1. 
However, assessing the embedding based on \dir-distance-based neighborhoods in the parameter space $\pspace$ yields a much lower \AUCe~of only 0.78. 
Computing \AUCg, i.e. computing neighborhoods using geodesic distances, in the parameter space -- recall that this is assumed to be the appropriate way to capture long-range distances on the manifold
-- leads to a further reduction, with an \AUCg~of only 0.553. This corresponds more closely to the visual impression, since \mds~is not able to unfold the intrinsic structure correctly. 

So we see that naive application of standard performance measures can indicate high quality embeddings even if the manifold is not accurately recovered at all -- at least if \textit{recovering} is defined as also \textit{unfolding} non-linear manifolds. The example also shows that the assessment of the quality of an embedding provided by these performance measures is highly sensitive to the choice of the distance metric used to determine the neighborhood structure in the space against which the embedding space neighborhoods are compared. 

These two pitfalls make the assessment of real data embeddings using surrogate performance measures particularly challenging since, in reality, the intrinsic structure is obviously unknown. In particular, automatic tuning approaches based on these measures have to be chosen and evaluated very carefully. 

\subsubsection{Evaluation of automatic parameter tuning}
To assess the overall approach of tuning and embedding functional data, we concentrate on the more challenging nonlinear settings \nlset. Since we want to assess the ability to detect and unfold intrinsic structure induced by a nonlinear parameter space here, identity warping is applied in all settings.
To thoroughly evaluate effects of the distance metric, we tuned each method on each setting based on \AUCe~and \AUCg, both based on the function space as well as the ground truth parameter space. That is, in total each method has been tuned four times for each setting.
Computing performance based on the parameter and the function space allows us to compare what is \textit{theoretically} possible -- based on the ground truth parameter space -- on the one hand, and what is \textit{practically} feasible -- based on the observable function space -- on the other hand. 
Reliable tuning and evaluation methods based on the functional data should provide similar answers and results as those achieved by tuning and evaluating on the true underlying parameter space.

\begin{figure}
    \center{A: Embeddings optimized via \AUCe in $\pspace$}
    \includegraphics[width=\textwidth]{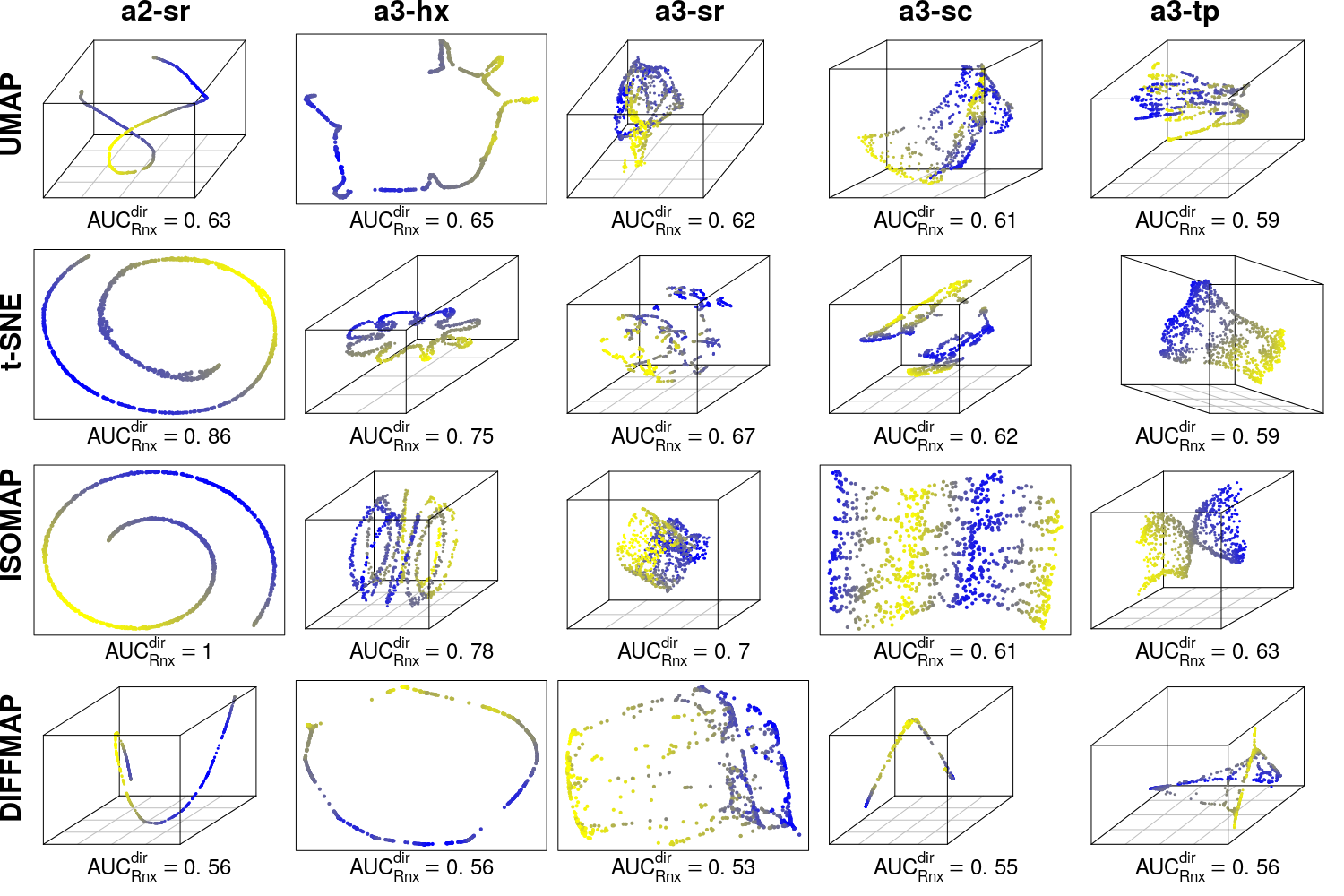}  \\
    
    \center{B: Embeddings optimized via \AUCg in $\pspace$}
    \includegraphics[width=\textwidth]{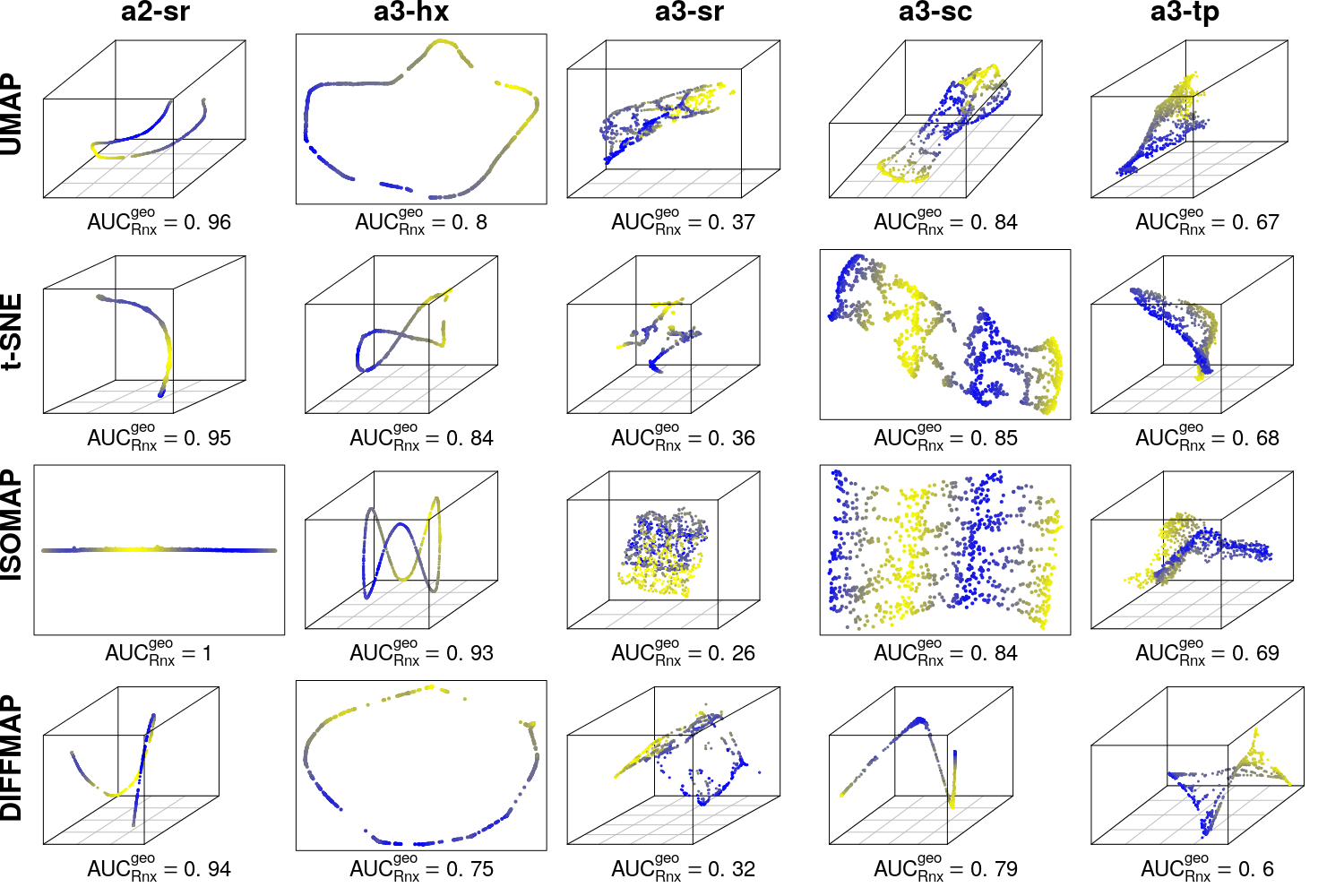}
    \caption{Parameter-space optimal embeddings of nonlinear settings \nlset. A: first four rows based on parameter space \AUCe-optimization. B: lower four rows based on \AUCg.
    Color scale encodes value of the first parameter in $\pspace$.} \label{fig:embs-nl-ps}
\end{figure}

\begin{figure}
    \center{A: Embeddings optimized via \AUCe in $\funspace$}
    \includegraphics[width=\textwidth]{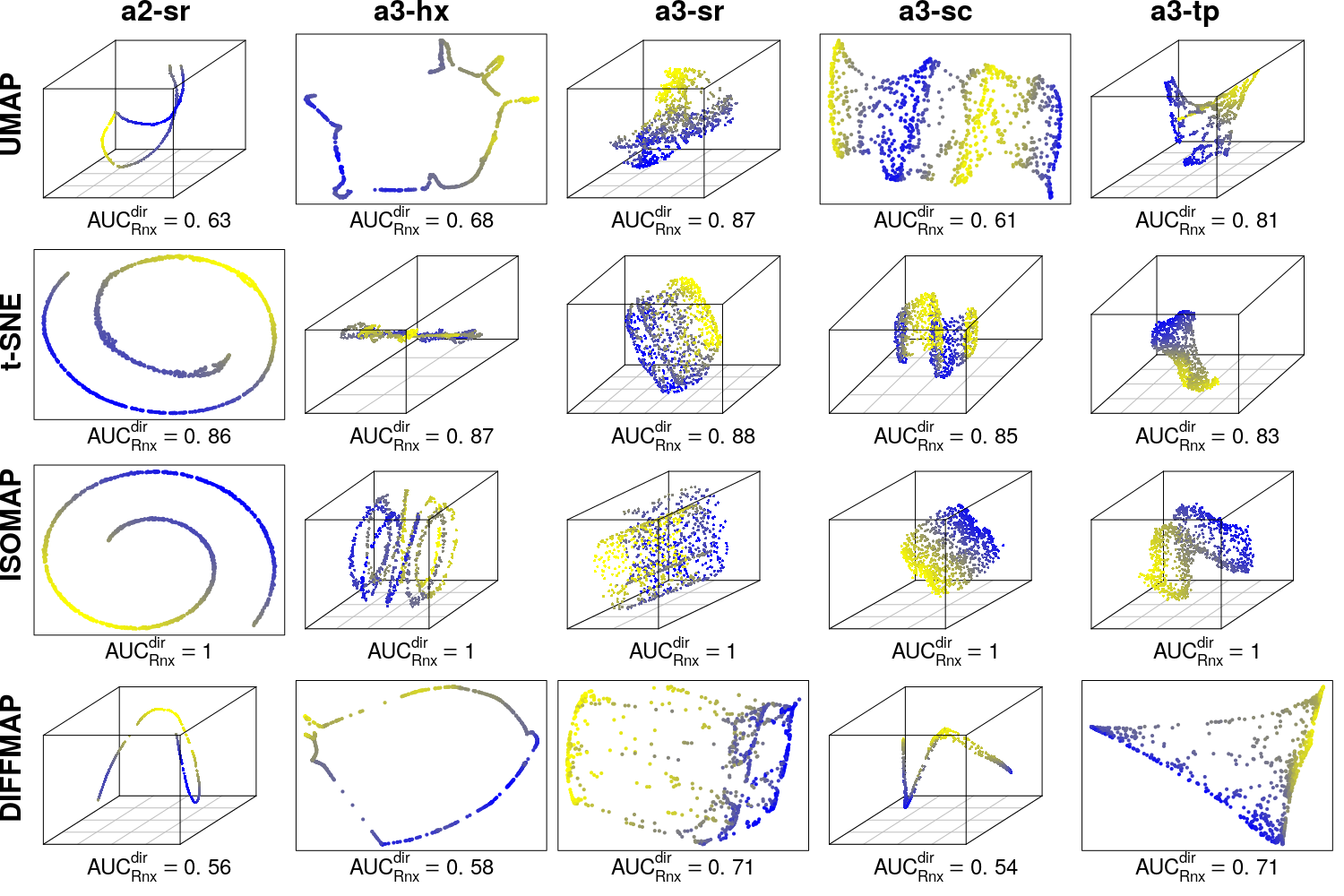}  \\
    
    \center{B: Embeddings optimized via \AUCg in $\funspace$}
    \includegraphics[width=\textwidth]{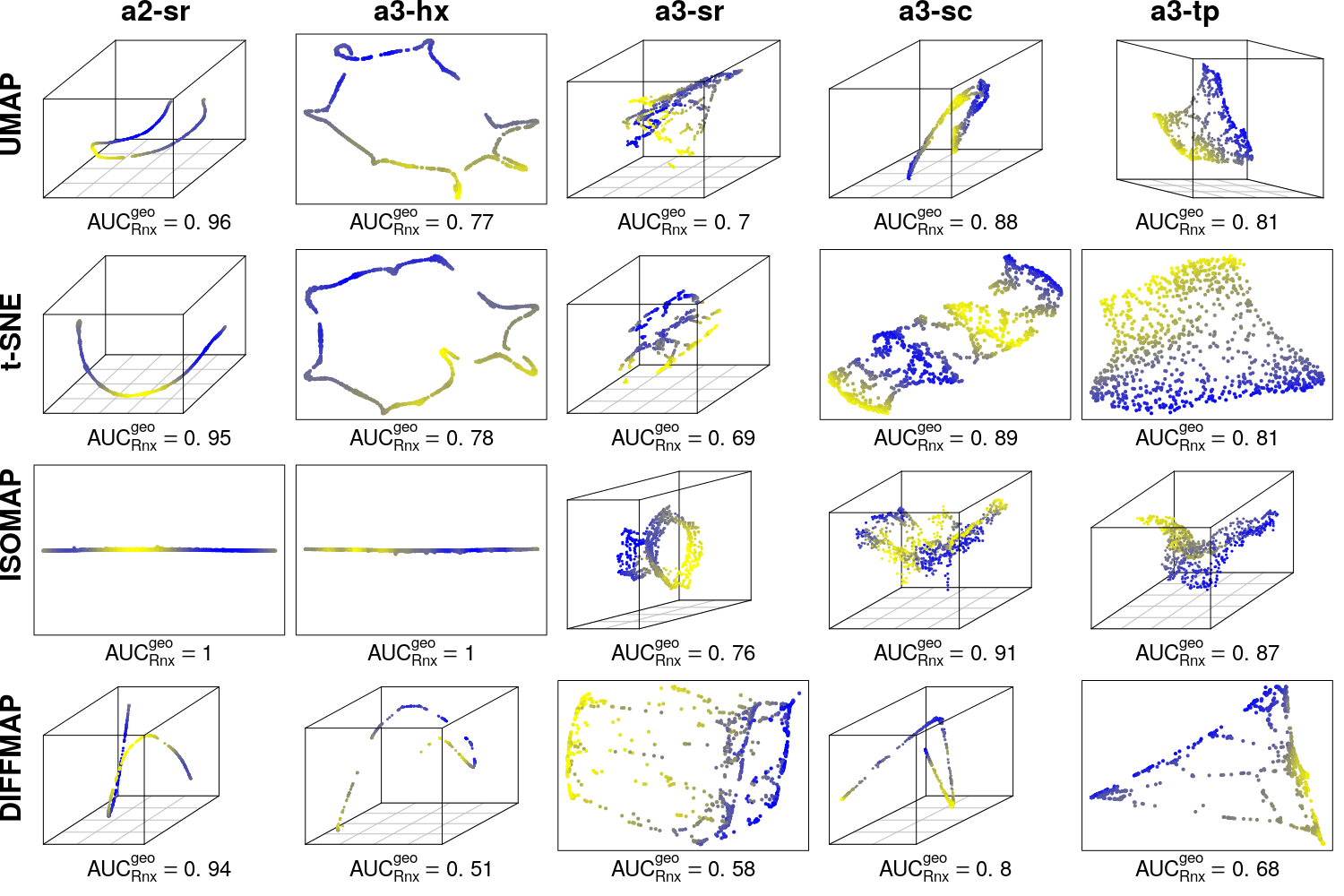}
    \caption{Functions-space optimal embeddings of nonlinear settings \nlset. A: first four rows based on function space \AUCe-optimization. B: lower four rows based on \AUCg. 
    Color scale encodes value of the first parameter in $\pspace$.} \label{fig:embs-nl-fs}
\end{figure}

Figures \ref{fig:embs-nl-ps} and \ref{fig:embs-nl-fs} display the resulting embeddings for the considered settings. The 
embeddings have been obtained via tuning based on the parameter space, i.e., maximizing agreement between parameter space neighborhoods and embedding space neighborhoods in Figure \ref{fig:embs-nl-ps} and obtained via tuning on the function space, i.e., maximizing agreement between function space neighborhoods and embedding space neighborhoods, in Figure \ref{fig:embs-nl-fs}.
Successful embeddings should unfold these data either to a circle (a3-hx, 2nd column) or to linear subspaces.
However, regarding the parameter space-optimized embeddings in Fig. \ref{fig:embs-nl-ps}, it becomes obvious that -- even if the true parameter space is used -- tuning can lead to embeddings that do not withstand visual inspection in that sense (e.g., see \tsne~and \imap~embeddings of a2-sr and a3-hx based on \AUCe; Fig. \ref{fig:embs-nl-ps} A, second and third row). For setting a2-sr, \AUCe indicates perfect embedding for \imap~and good embedding for \tsne. The corresponding embeddings based on \AUCg (Fig. \ref{fig:embs-nl-ps} B, second and third row), however, withstand visual inspection far better. This already indicates that tuning based on geodesic distances can lead to better results than simply relying on direct distances. 
Turning to the function space optimized embeddings (Fig. \ref{fig:embs-nl-fs}), we see that things can get even more involved in reality. Consider, for example, the \imap~embeddings again. 
The embeddings based on \AUCe~(Fig. \ref{fig:embs-nl-fs} A) for a3-sc as well as for settings a2-sr, a3-hx and a3-sr are not satisfactory, even though high performances are indicated by the measure. 

These discrepancies between measured and actual performance are due to the fact that, in the case of direct distances, the performance measure is based on a suboptimal distance metric in the high-dimensional spaces $\pspace$ and $\funspace$. For strongly nonlinear settings, direct distances seem to be insufficient for tuning methods so that they correctly reflect the intrinsic structure. Analogously, this applies to the function space as well, since the \ltwo~metric in the function space is structurally very similar to Euclidean distance in a Euclidean space. So \ltwo-based neighborhoods are likely to be different from neighborhoods based on \textit{geodesic} distances in the function space whenever the functional manifold is nonlinear. 
Due to the more complex structure of the function space the effect seems to be intensified (for example, see \imap~embeddings for a3-sc: based on the parameter space, direct distances were sufficient to recover the manifold, whereas in the function space the intrinsic structure was only recovered if tuned based on geodesic distances). Next to the effects of nonlinear domain warping this is another example for the specific challenges of nonlinear dimension reduction in FDA settings.  

Turning to the remaining methods, the picture is a little more difficult to make sense of, because the embeddings do not yield such clear differences as \imap~and \tsne. 
In general, \dmap~and \umap~show arguably better embedding results based on \AUCg (e.g. a2-sr, a3-hx, Fig. \ref{fig:embs-nl-ps}), but, on the other hand, they benefit less from using geodesic distances for tuning (e.g. a3-sc, a3-hx, a2-sr, Fig. \ref{fig:embs-nl-fs}). \dmap~embeddings, in particular, differ the least among \AUCe~and \AUCg. 
Moreover, in some cases the underlying manifold is hardly recognizable or not successfully unfolded, in particular this holds for the \dmap~embeddings in settings a3-sr, a3-sc and a3-tp.

\begin{figure}
    \includegraphics[width=\textwidth]{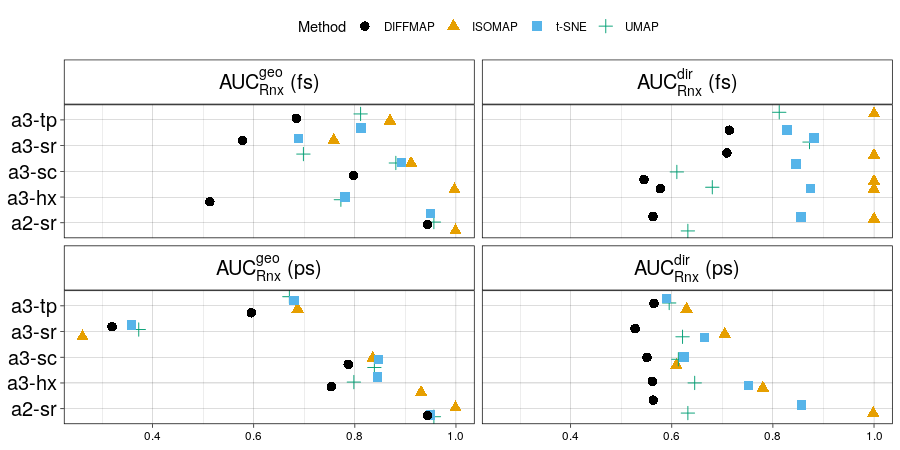}
    \caption{Function (fs) and parameter (ps) space optimal \AUC values based on geodesic and \dir~distances for settings a2-sr, a3-hx, a3-sr, a3-sc, a3-tp.} \label{fig:dotplot}
\end{figure}

To quantify the differences between using geodesic rather than \dir~distances to optimize and compute performance measures, Figure \ref{fig:dotplot} shows the different optimal \AUC-values for all nonlinear settings obtained on the function space and the parameter space. The best values achieved based on the function space differ strongly from the ones based on the parameter space in several cases. In general, optimization via \AUCg~leads to smaller differences between tuning on function and parameter spaces distances. 
This is also reflected in Table \ref{tab:l2-vs-geo}, which shows the absolute differences $\Delta_{\bar{a}}^{m} := |\bar{a}_{ps}^m - \bar{a}_{fs}^m|$, with $\bar{a}_{ps}^m$ and $\bar{a}_{fs}^m$ the mean optimal \AUC~based on parameter respectively function space. 
The mean values $\bar{a}_{ps}^m$ and $\bar{a}_{fs}^m$ are computed over the \imap, \dmap, \tsne, and \umap~embeddings and the settings a2-sr, a3-hx, a3-sc, a3-tp. Since setting a3-sr could not be embedded successfully with any of the methods even if optimized over the parameter space, it is excluded. Clearly, optimal \AUCg~(values based on the geodesic distances) differ less between function space and parameter space than \AUCe~(values based on the \dir~distances) for \imap~and \tsne, while there is not much of a difference for \umap~and \dmap. This is in line with the visual impression that \imap~and \tsne~yield clearly better embeddings based on tuning via \AUCg~for these settings.\\
This also indicates that it is frequently more appropriate to use geodesic distances -- especially in function spaces -- for performance assessment and tuning. 

\begin{table}
    \caption{Comparing performance assessment based on geodesic and \dir~distances using absolute difference of mean \AUC~in parameter space and function space. 
    Setting a3-sr is excluded because no visually appropriate embeddings could be obtained.}
    \centering
    \begin{tabular}{|l|l|l|}
    \hline
    Method & $\Delta_{\bar{a}}^{dir}$ & $\Delta_{\bar{a}}^{geo}$\\
    \hline
    \imap & 0.246 & 0.081 \\
    \tsne & 0.146 & 0.060 \\
    \umap & 0.064 & 0.052 \\
    \dmap & 0.085 & 0.043 \\
    \hline
    \end{tabular}
    \label{tab:l2-vs-geo}
\end{table}

To sum up, we have seen that it is possible to obtain successful embeddings for functional data and that automatic parameter tuning can be applied in these simulation settings. Moreover, using distances in function space seems to be a reliable alternative to using distances in ground truth parameter space. In some of the complex settings \nlset, however, tuning is successful only if based on geodesic distances rather than direct distances to define the neighborhoods in the high-dimensional space. 
In general, these are promising results indicating that (automatically) obtaining high quality embeddings for real functional data is feasible.
Yet, the approaches should not be applied lightly to real data. As we have seen, some of the methods may yield suboptimal or nonconforming embeddings even if tuned properly. Moreover, not every setting seems to be amenable to a successful embedding (e.g. see setting a3-sr) -- factors which can lead to multiple nonconforming or misleading embeddings and overoptimistic or invalid conclusions if not assessed carefully.

\section{Real data application}\label{sec:real-data}

We now turn to real data examples in this section in order to verify the practical utility of the insights from our simulation study.
As motivated in Section \ref{sec:background:study-design}, we first apply our approach to two settings where the intrinsic structure of the data can be inferred from domain knowledge to a certain extent. Subsequently, we investigate a fully unsupervised real data example with completely unknown structure. In addition to \AUC, we also evaluate embedding performance via \qlocal~here in order to investigate the distinctions between local and global performance measures. To begin with, we give a short description of all three date sets.

\subsection{Data sets}

We apply the embedding methods on two functional and one image data set. For two data sets, the COIL data and the earthquake data, the intrinsic structure can be inferred from prior knowledge in advance, at least to a certain extent. 
The intrinsic structure of the third data set, a spectrography data set, is not known.
Figure \ref{fig:realdat} shows example observations of the three real data sets. More details are given in the sections devoted to the specific data set.

\subsubsection{COIL data}\label{sec:real-desc-coil}
COIL20 \cite{coil20} is an image data set consisting of 128 x 128 pixel images of 20 objects. We use this data set albeit it is not functional, because it is a real data set for which the intrinsic structure can be inferred from substantial considerations and is nonlinear. For each object, 72 pictures were obtained by rotating the object around itself and taking a picture every 5 degrees of rotation. The end position equals the starting position and each picture reflects the object at a different angle. Thus, for each object the COIL20 data set contains 72 observations with 16384 features containing single pixel intensities. 

\begin{figure}
    \includegraphics[width=\textwidth,height=4cm]{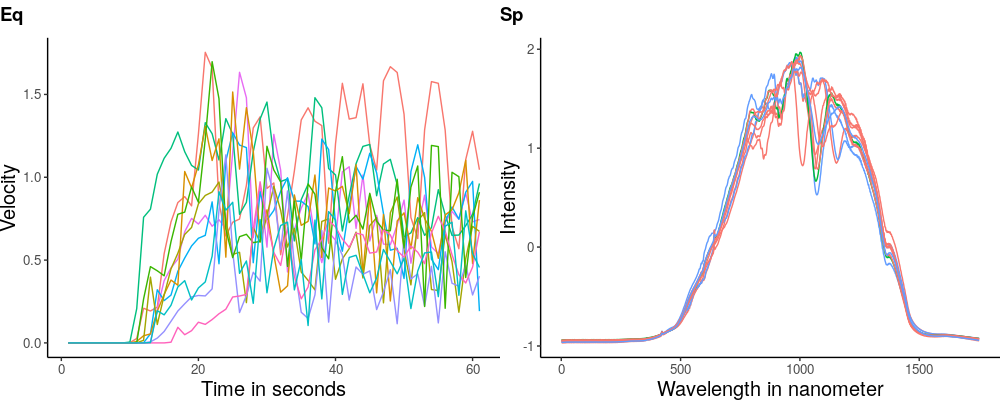}
    \includegraphics[width=\textwidth,height=4cm]{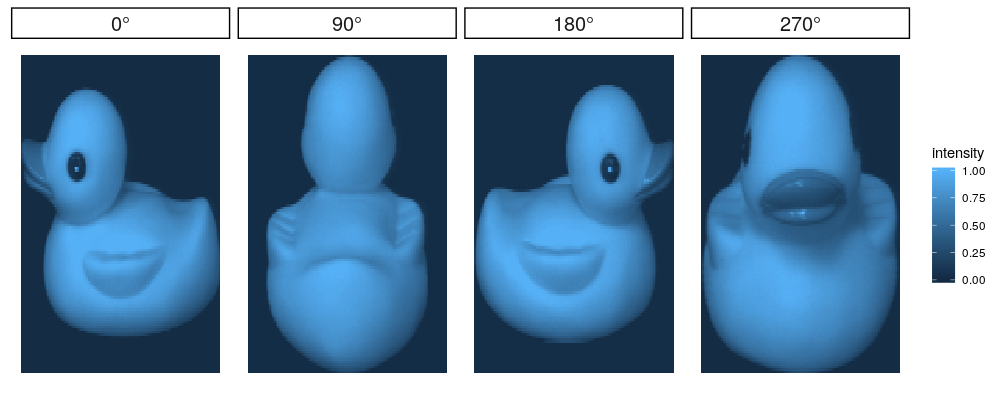}
    \caption{Upper row: Ten example observations of each the earthquake data (Eq) and the spectrography data (Sp). Lower row: Four images of the COIL object.} \label{fig:realdat}
\end{figure}

For this study we use a subset of the COIL20 data containing only the pictures of the first object as the high-dimensional data to be embedded. This means, the data set we use contains 72 pictures of the same object depicted in Figure \ref{fig:realdat} at different angles ranging from 0 to 360 degrees. Considering this setup, the 5-degree-picture should -- for example -- have approximately the same distance to the 0-degree-picture as the 355-degree-picture. The intrinsic structure of the data set is thus expected to be circular and one dimensional, i.e., a setting supposed to be comparable to setting a3-hx. 

Aside from these insights -- which can be inferred from the original description of the data generating design and applies to all of the 20 COIL objects -- for the specific object regarded here further considerations can be made, if one closely examines Figure \ref{fig:realdat}. Due to the axial symmetry of the object, it appears to be more similar to itself at positions 0 and 180 degrees than at 90 and 270 degrees. This can be an indication of further existing structure which might be present in this specific example, but whether and how this is reflected in the embeddings is difficult to assess.

\subsubsection{Earthquake data}
The second real data set contains functional data of a seismological \textit{in silico} experiment with both phase and amplitude variation described in \cite{happ_general_2019}. It contains 1558 observations observed on 61 grid points. Each observation represents 60 seconds of absolute ground movement velocities at a virtual seismometer location for a simulated earthquake. The original investigation based on multivariate functional PCA of phase and amplitude variation revealed a two dimensional linear structure of the data \cite{happ_general_2019} reflecting the spatial distances of the virtual seismometers to each other and to the simulated earthquake's hypocenter. That is, from the analyses of the previous study we can infer that this is a data set with phase and amplitude variation and -- following our framework -- supposedly a relatively simple (linear) underlying parameter space, so a setting assumed to be comparable to settings i2-l or a2-l, for example.

\subsubsection{Spectrography data}
Finally, we consider another functional data set with 1004 functional observations observed on a grid of length 1751. The data was originally generated to investigate how forged spirits can be detected noninvasively via vibrational spectroscopy of the ethanol level, i.e. each observation is a spectrograph based on 1751 different wavelengths (see \cite{large2018} for more details on the data set). The data is usually used as a classification problem and can be obtained\footnote{\url{http://www.timeseriesclassification.com/description.php?Dataset=EthanolLevel}} separated into a training set with 504 observations and a test set with 500 observations. Since we are in an unsupervised setting, we merged the training and test set and use the joined data set as the high-dimensional data to be embedded. Here, we cannot make any justifiable assumptions about the intrinsic structure and it is unclear what a successful embedding should look like.

\subsection{Application to real data with known structure}
\begin{figure}
    \center{Earthquake data}
    \includegraphics[width=\textwidth]{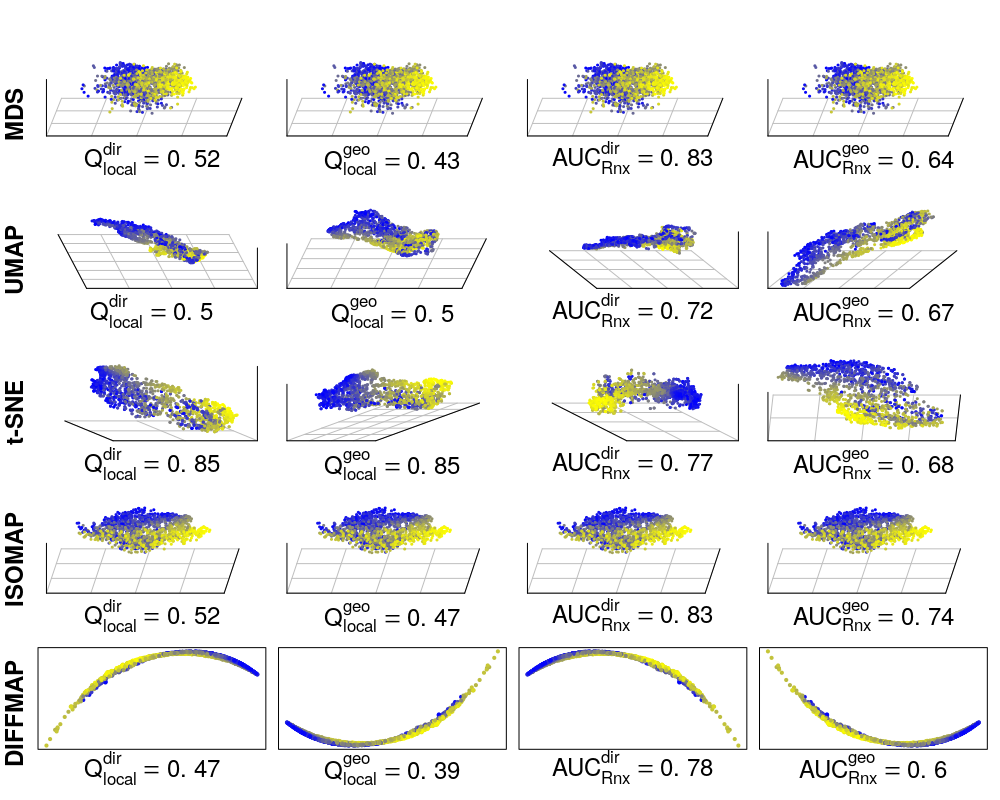}
    \caption{Embeddings of the earthquake data. First two columns obtained by optimization via \qlocal, latter two columns via \AUC. Color scale encodes distance to hypocenter.} \label{fig:embs-eq}
\end{figure}

\begin{figure}
    \center{COIL data}
    \includegraphics[width=\textwidth]{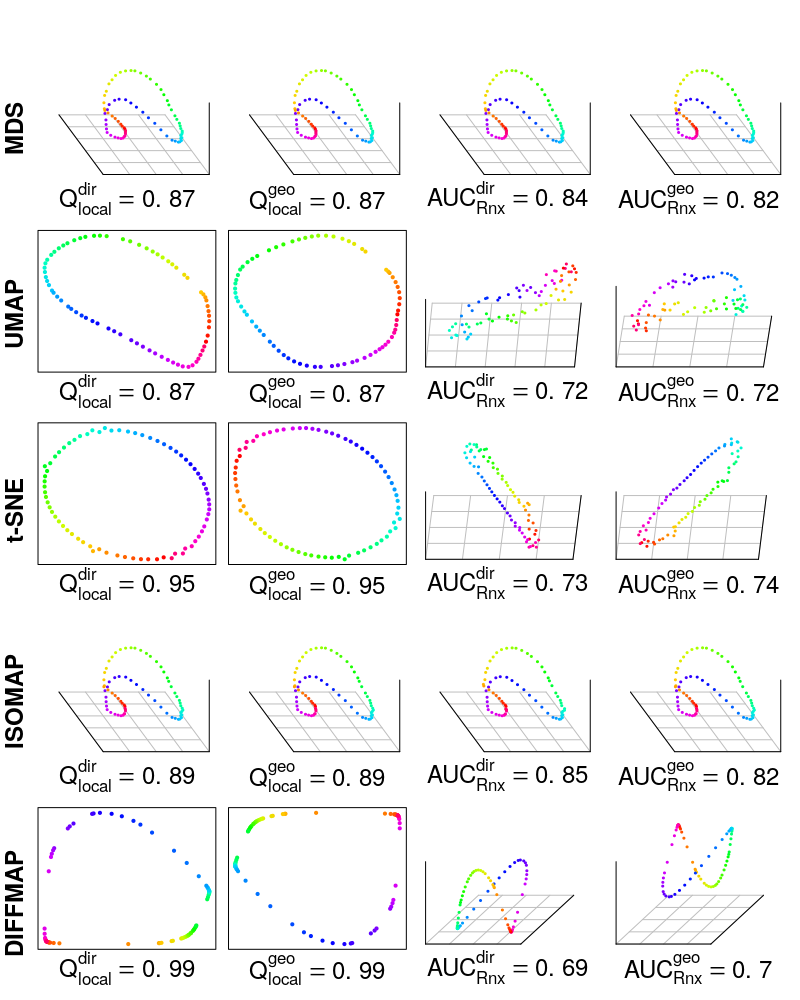}
    \caption{Embeddings of the COIL data. First two columns obtained by optimization via \qlocal, latter two columns via \AUC. Color scale encodes rotation angle.} \label{fig:embs-coil}
\end{figure}

Figure \ref{fig:embs-eq} displays the embeddings for the earthquake data and Figure \ref{fig:embs-coil} for the COIL data. The first two columns for each data set are obtained by tuning via the local performance measure \qlocal, while the latter two columns are obtained by tuning via the global performance measure \AUC.

Most importantly, the results show that what has been observed for the simulated data also holds for real data: the low dimensional intrinsic structure -- both linear and nonlinear -- can be successfully embedded. Several embeddings of the COIL data show a 1-dimensional, circular structure, while a 2-dimensional linear structure results for the earthquake data. Moreover, closer examination reveals further interesting insights. 

First of all, considering the earthquake data it appears that there is not much of a difference between embeddings based on \AUCg~and \AUCe~in a setting with nearly linear intrinsic structure. The \tsne~and \umap~embeddings based on \AUCe~visually appear to be inferior to the ones based on \AUCg. Based on the performance values all embeddings based on \AUCg~perform somewhat worse. But in general the performance differences are negligible. This is a desirable result in line with theoretical considerations: On approximately linear subspaces, geodesic distances ought to resemble direct distances. 

Next to these findings, the COIL data opens up a rich pool of interesting insights that extend the understanding of embedding methods beyond that obtained in the simulation study.
First of all, as outlined in Section \ref{sec:real-desc-coil}, a 1-dimensional circular structure can be assumed from the data generating procedure and this structure is detected by four of the embeddings. However, due to the symmetries of the rotating object, additional structure could be assumed and this additional structure seems to be recovered in several embeddings as well -- in the case of some of the methods only if the global performance measure \AUC~is used for tuning, however.  Consider the \tsne~embeddings where the effect is most prominent. In the two rightmost columns tuned for \AUC, the structure is ellipsoid, while it is circular in the two leftmost columns tuned for \qlocal. These nonconforming embeddings can be explained with the axial symmetry of the object. Due to the symmetry, the object is more equal to itself at position 0 and 180 degrees than at 90 and 270 degrees, which is a \textit{global} characteristic (locally, i.e. within a small range of rotation angles, the objects looks similar to itself everywhere). The local performance measure \qlocal~is not able to reflect this global characteristic of the data sufficiently and the aspect is lost in \umap, \tsne, and \dmap~embeddings if tuned based on \qlocal. 
These examples demonstrate that global structural properties can easily be ``lost in translation'' if an embedding is not tuned properly. It also has to be emphasized that -- in contrast to \tsne~-- those \umap~embeddings which sufficiently preserve global structure do not simultaneously preserve local structure in this setting.
The situation is a little different for \imap. As can bee seen in Figure \ref{fig:embs-coil}, the global structure is recovered in the embeddings, both based on \qlocal~and \AUC. However, an additional dimension is needed to reflect this in the embedding, since all embeddings result in a twisted ring with an upward bend at two positions opposed to each other. That is, \imap~is not able to fully recover the structure in this example in the lowest possible number of embedding dimensions (similiar to \dmap).
Considering the \mds~embeddings of the COIL data, we see that they are almost completely equal to the \imap~embeddings (only the performance indicated by the values of \qlocal~is slightly worse for \mds). A possible explanation is that due to the low amount of observations (only 72 for COIL), the geodesic distances do not differ sufficiently from the direct distances and \imap~basically reflects \mds~embeddings. This is supported by a couple of insights which can be gained by contrasting these results with the results of the simulations study. \\
\indent First of all, regarding the \mds~embeddings of the COIL data and the a3-hx data, we see that in both situations the intrinsic structure (a twisted ring and the helix, respectively) is recovered in principle. Yet, while the intrinsic structure of a3-hx gets unfolded by \imap, the same is not true for COIL. The most fundamental difference between these two examples is the number of observations (1000 for a3-hx, 72 for COIL), which points towards the conclusion that, based on the low number of observations, a sufficient shortest path graph cannot be constructed for the COIL data such that \mds~would benefit from it compared to simply using direct distances. Moreover, additional experiments in which the number of observations was increased from 1000 to 5000 in setting a3-sr showed that increasing the number of observations can lead to embeddings which successfully unroll the manifold -- the failure of the methods to do so in our original experiment in setting a3-sr is likely to be due to too few observations. Finally, the conclusion is also supported by the fact that there are in general almost no differences between the COIL embeddings tuned via \AUCe and \AUCg -- neither visually nor quantitatively. Recall, however, that in the simulated settings there have been large differences and -- despite the fact that \tsne~and \umap~are able to recover the global structure here -- it was frequently crucial to use geodesic distances so that the intrinsic structure could be unfolded. 

To sum up, the investigation of these two examples confirms that low dimensional intrinsic structure of real (functional) data can be automatically and successfully embedded. However, some pitfalls were clearly identified. For one, certain structural properties of the functional manifold can easily get lost in the embedding if it is not tuned with the correct strategy, which makes the assessment of fully unsupervised settings specifically challenging. Secondly, we saw that using geodesic instead of direct distances is only beneficial if the number of observations is sufficiently large in a nonlinear setting. Regardless, their use does not seem to be harmful even in settings with few observations and/or settings with linear structure of the functional manifold.

\subsection{Application to real data with unknown structure}
So far, we have seen that in several settings -- simulated as well as real -- where the intrinsic structure is known at least to certain extent, functional data can be  embedded successfully and that automatic tuning can be used to obtain suitably faithful embeddings. On the other hand, we identified some specific pitfalls. 
In this subsection, we illustrate the resulting challenges of nonconforming embeddings with a fully unsupervised real data example and point out an approach possibly allowing to gain some further insights in such fully unsupervised settings. 

\begin{figure}
    \center{Spectrography data}
    \includegraphics[width=\textwidth]{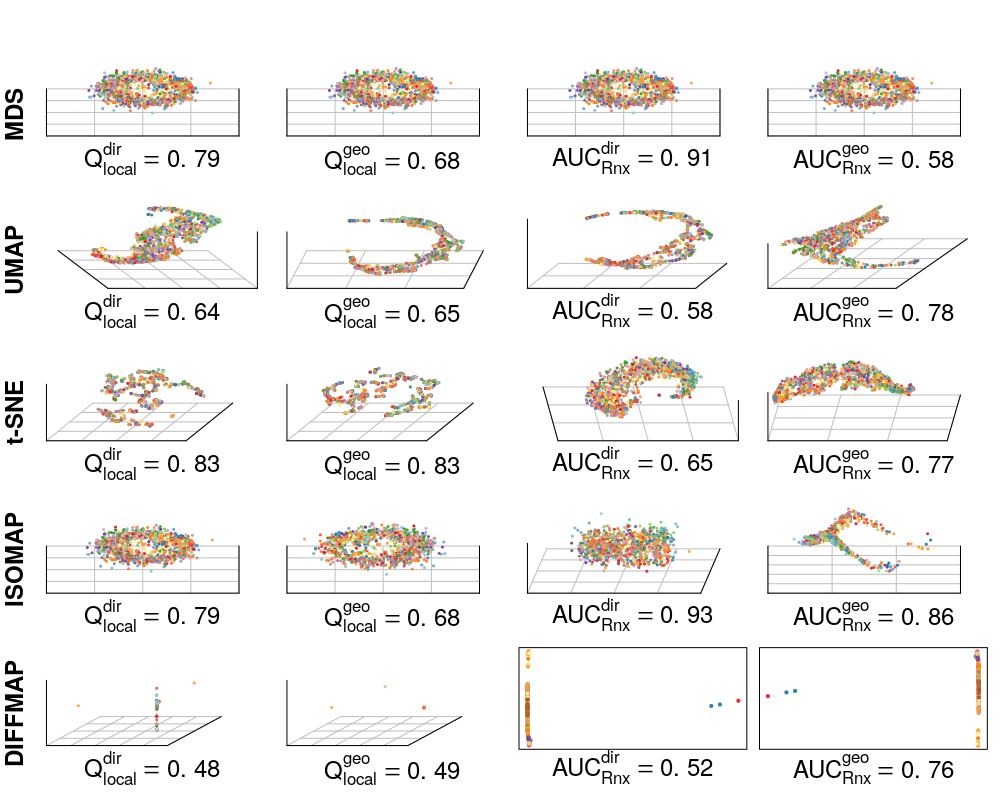}
    \caption{Embeddings of the spectrography data. First two columns obtained by optimization via \qlocal, latter two via \AUC. To improve visual differentiation, points are colored according to their observation number.} \label{fig:embs-sp}
\end{figure}

The embeddings of the spectrography data are depicted in Figure \ref{fig:embs-sp}. The major problem is that there are -- overall -- two nonconforming structures which are detected. On the one hand, a closed, circular 3-dimensional structure -- a ``donut" --, detected by \mds, all \imap~embeddings except the one tuned via \AUCg, and arguably also the \tsne~embeddings based on \qlocal. On the other hand, there is a curved, open 3-dimensional structure detected by \imap~based on \AUCg, \tsne~based on \AUCg~and to lesser extend \AUCe, and \umap. In this example, it is not possible to decide which of the embeddings better describes the true structure by visual inspection or reference to prior knowledge, nor is it expedient to simply maximize performance measures. E.g., \AUC~is similarly high for \imap~both based on geodesic as well as direct distances, yet they lead to nonconforming embeddings. The drawbacks and pitfalls of the performance measures described in the simulation study are fully apparent here. Although performance measures are available, deciding between nonconforming embeddings is far from straight-forward.  

However, the results gathered so far allow us to introduce additional decision criteria: 
First of all, we saw that, in the COIL and the a3-hx examples, closed structures were detected and recovered by all methods irrespective of the performance measures used for tuning. That is, in a setting with a not ``fully'' unfoldable structure which is closed in some way or the other, this structure was recovered in all cases considered. Specifically, that included embeddings obtained with \AUCg. On the other hand, if a nonlinear structure is `fully' unfoldable, we saw that, in several cases, a fully unfolded embedding was achieved only if the embeddings were based on \AUCg (e.g. see a3-sc) -- provided there was enough data.
Considering the spectrography data, we see that none of the embeddings based on \AUCg~-- except MDS of course -- show the closed circular structure. 
It may thus not be too far-fetched to infer that, if the intrinsic structure of this data set were closed and circular in reality, this would also be reflected by at least some of the embeddings based on \AUCg~(as was observed for example for the COIL and the a3-hx data) and that the ``donut" does not actually resemble the true intrinsic structure sufficiently. In fact, one possible explanation could be that direct distances might falsely indicate that two objects at opposite ends of an open, but curved manifold  are close together by taking a shortcut ``through" the ambient space, and thus connect these parts of the functional manifold in the embedding space. Similarly, based on local measures -- that is, from a local perspective -- such points might also appear close and global characteristics cannot be reflected sufficiently as was the case in the COIL example. This might result in unconnected manifold regions being pulled together, which might be appropriate locally, but yields a wrong impression from a global perspective. The fact that \umap, a method which is mainly concerned with an accurate representation of the local structure \cite{mcinnes_umap_2018}, does not show a closed structure even if based on \qlocal, contributes to this conclusion. Note, however, that other explanations might be possible. For example, \mds~embeddings reflected the true intrinsic structure in the simulated settings, which may also be the case here and the other, ``unclosed" structure might result from undiscovered effects of the approaches. Nevertheless, this example points into a direction to gain insights in settings where so far no judgment is possible.

\section{Discussion}\label{sec:discussion}
Our results indicate that nonlinear dimension reduction methods can detect and ``unfold" the manifold structure of functional data. For our settings, \imap~and \tsne~were seen to be particularly successful methods. Based on the same tuning regime, the other methods under comparison -- \dmap, \umap, and \mds~-- frequently did not obtain similarly useful embeddings. However, we focused on recovery of the global data manifold structure in the embedding space.
This is likely to affect \umap~adversely since it is optimized for high-fidelity reconstruction of local structures. The conclusion should thus not be that \tsne~or \imap~are superior for embedding functional data in general. Similar remarks apply for \dmap, which is also a local method and performed less well than \umap~in our experiments. Especially in higher-dimensional real data settings, it frequently led to degenerate embeddings. 

Furthermore, tuning strategies based on surrogate performance measures such as \AUC~should be applied with caution, since they may lead to very misleading embeddings with far from optimal configurations.
In fact, we found evidence that embedding performance strongly depends on the distance metric $m$ supplied to the performance measure used for tuning. Our results suggest that the use of geodesic distances -- in particular in function space -- is more likely to yield suitable embeddings if faithful representation of global structure is of importance: tuning embeddings based on the functional geodesic distances rather than direct distances yielded embeddings that were frequently much more similar to the ground truth structure in the simulation study and the expected structure in the real data examples.

Taking all insights obtained in this study into account, we propose to use the following nuanced approach to achieve more reliable embeddings of functional data: (1) Embeddings should be computed automatically by the described tuning approach. (2) At least two embedding methods should be included to account for different method performances. In addition, a tuning-free reference method should be included, for which we suggest \mds, since it can recover intrinsic structures in simple cases although it does not ``unfold'' them. (3) Embeddings should be computed based on optimizing a local as well as a global performance measure. (4) Geodesic distances should be used. Tuning based on geodesic distances worked better for some methods, specifically in complex nonlinear settings and did not have adverse effects on performance in any other settings. In addition, discrepancies between geodesic-based embeddings and direct-distance-based respectively local-performance-based embeddings can provide clues on the likely complexity of the intrinsic structure (closed vs. nonclosed, linear vs. nonlinear).

Moreover, some general questions are raised as well, as the last aspect strongly affects how to appropriately tackle specific unsupervised problems with manifold methods. For problems such as clustering or outlier detection, where preserving local structure can be sufficient, relying on non-geodesic distances can be appropriate. Yet, according to our findings, this can not simply be transferred to other tasks where global structure is more important, for example using manifold methods as a preprocessing or feature engineering step. In this setting, \textit{unfolding} the manifold, i.e., detecting and simplifying the global structure, is important because reliable low dimensional representations not only improve visualization and exploration of functional data, the embedding coordinates can also be exploited as features, which preserve the essential information contained in the functional data, for supervised learning (i.e., modeling and inference) tasks. Using direct distances to assess embeddings quantitatively or to optimize learned embeddings in an automated fashion is (more) likely to lead to misleading results in this context.

In summary, our results show the potential of extending manifold methods for NDR to function-valued data, but also
reveal  challenges that are likely to come up in applications.
In order to achieve reliable low-dimensional representations of functional data for visualization and exploration or to serve as feature inputs in supervised learning tasks, these issues will require additional attention by the research community. In future work, we will investigate the effect of noise-corrupted observations on the estimation of geodesic distances for functional data, since errors that shift observed functions off the functional manifold are likely to affect the recovery of geodesic distances adversely. In addition, the effects of grid resolution and data set size as well as the definition of alternative distance metrics which specifically account for certain characteristics of functions -- for example, separate amplitude and phase distances \cite{srivastava_registration_2011} -- are further important aspects.

More generally, this study should be considered in the light of a growing debate on replicability in methodological research. As has been outlined by many \cite[e.g.]{kobak_umap_2019,lucic2018gans, boulesteix2020replication,hutson2018artificial}, methodological research 
claiming to show superior performance of its proposed methods and approaches in one way or another can frequently not be confirmed in independent replications. Our aim in this work was to provide a fairly neutral evaluation by design, pointing out specific pitfalls and drawbacks of several widely used manifold learning algorithms and possible meta learning methods on a wide range of functional data settings. However, other evaluation frameworks are certainly possible and may yield additional insights and qualifications of our conclusions. 
As long as there is no general benchmarking and evaluation regime generally agreed upon, results of all such studies will depend on the choice of this framework to some extend. In that regard, our study is also intended to serve as a starting point and we hope that it may contribute to initiate a discussion -- similar to the ones in supervised learning and cluster analysis -- on how to conduct neutral evaluations and foster replicable results in the important field of NDR and manifold learning.

\section*{Technical details}
All code necessary to reproduce the experiment can be found on GitHub \url{https://github.com/HerrMo/fda-ndr}.
To conduct the experiments we used R 3.6.3 \cite{R-core} on a system with Linux Mint Cinnamon 19.2 and R packages \verb|vegan| \cite{R-vegan} for ISOMAP, \verb|diffusionMap| \cite{R-dmap} for DIFFMAP, \verb|Rtsne| \cite{R-rtsne} for t-SNE, and \verb|umap| \cite{R-umap} for UMAP. For the performance measures we used code of the functions \verb|auc_rnx| and \verb|q_local| from the \verb|dimRed| package \cite{kraemer_dimred_2018}. To compute \mds~we used \verb|cmdscale| of package \verb|stats| and \verb|isomapdist| of package \verb|vegan| to compute geodesic distances.

\section*{Acknowledgments} This work has been funded by the German Federal Ministry of Education and Research (BMBF) under Grant No. 01IS18036A. The authors of this work take full responsibilities for its content.

\section*{Declarations of interest} None.

\end{document}